\pdfoutput=1
\documentclass[10pt,twocolumn,letterpaper]{article}

\usepackage[pagenumbers]{cvpr} 

\usepackage{graphicx}
\usepackage{amsmath}
\usepackage{amssymb}
\usepackage{booktabs}
\usepackage{graphicx}
\usepackage{xcolor}
\usepackage{stfloats}
\usepackage{comment}
\usepackage{graphicx}
\usepackage{wrapfig}
\usepackage{caption}

\usepackage[pagebackref,breaklinks,colorlinks]{hyperref}

\usepackage[capitalize]{cleveref}
\crefname{section}{Sec.}{Secs.}
\Crefname{section}{Section}{Sections}
\Crefname{table}{Table}{Tables}
\crefname{table}{Tab.}{Tabs.}

\makeatletter
\newcommand{\thickhline}{
    \noalign {\ifnum 0=`}\fi \hrule height 1pt
    \futurelet \reserved@a \@xhline
}
\makeatother

\begin{document}

\title{Neural Poisson: Indicator Functions for Neural Fields}

\author{
	Angela Dai \hspace{2cm} Matthias Nie{\ss}ner \vspace{0.1cm} \\
	Technical University of Munich \\
}

\twocolumn[{%
	\renewcommand\twocolumn[1][]{#1}%
	\maketitle
	\begin{center}
		\includegraphics[width=\linewidth]{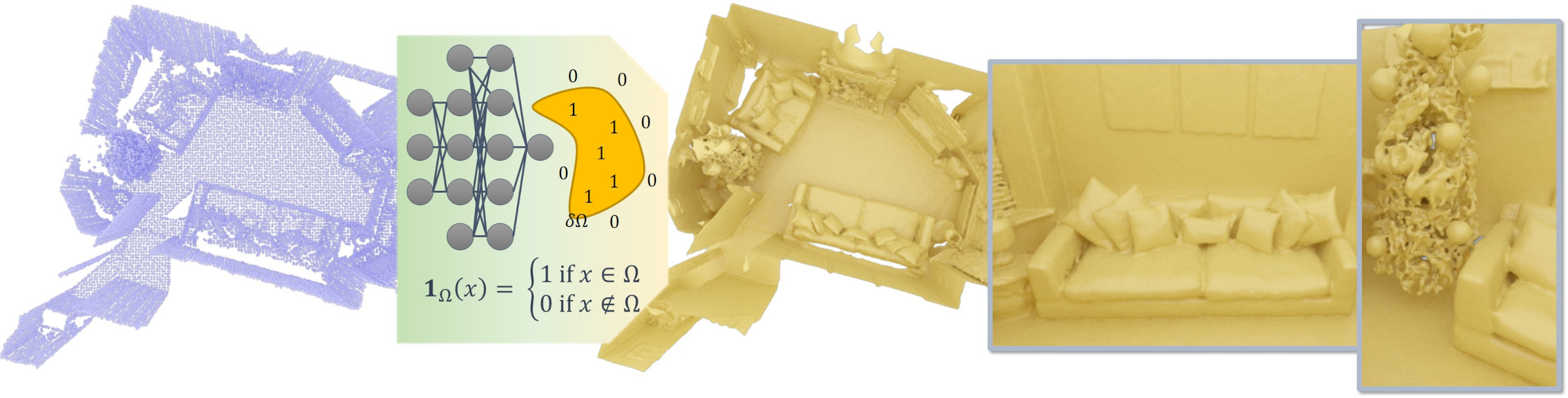}
		\captionof{figure}{
		We propose an indicator-based formulation for 3D reconstruction with neural fields. By using neural fields to approximate an indicator function, we can easily incorporate inherent empty space observations from real scan scenarios to produce high-fidelity, accurate surface reconstructions.
		}
		\label{fig:teaser}
	\end{center}    
}]

\maketitle


\begin{abstract}
   Implicit neural field generating signed distance field representations (SDFs) of 3D shapes have shown remarkable progress in 3D shape reconstruction and generation.
   We introduce a new paradigm for neural field representations of 3D scenes; rather than characterizing surfaces as SDFs, we propose a Poisson-inspired characterization for surfaces as indicator functions optimized by neural fields.
   Crucially, for reconstruction of real scan data, the indicator function representation enables simple and effective constraints based on common range sensing inputs, which indicate empty space based on line of sight.
   Such empty space information is intrinsic to the scanning process, and incorporating this knowledge enables more accurate surface reconstruction.
   We show that our approach demonstrates state-of-the-art reconstruction performance on both synthetic and real scanned 3D scene data, with $9.5\%$ improvement in Chamfer distance over state of the art.
\end{abstract}

\section{Introduction}

Surface reconstruction has been a fundamental problem in computer vision and graphics, with wide applications to robotics, mixed reality, and content creation.
A continuous surface reconstruction enables effective physical reasoning for robot navigation and interaction, with a dense geometric representation enabling 3D-based semantic inference.
Surface geometry representations also fit directly into content creation pipelines, enabling capture of 3D assets from real-world environments.
In particular, the ease of availability in acquiring sparse depth measurements of real environments is widespread, e.g., from multi-view images using structure-from-motion pipelines such as COLMAP~\cite{schoenberger2016sfm,schoenberger2016mvs}, or range-sensing devices such as the iPhone or Microsoft Kinect.
This has driven the need for high-fidelity surface reconstruction from sparse point measurements.

Recently, advances in neural fields have shown remarkable progress in representing 3D shape surfaces \cite{park2019deepsdf,mescheder2019occupancy,gropp2020implicit,sitzmann2020implicit}, by optimizing for functions $f$ such that 
\begin{equation*}
    \{\mathbf{x}\in \mathbb{R}^3 | f(\mathbf{x},\theta) = 0\},
\end{equation*}
where $f:\mathbb{R}^3 \times  \mathbb{R}^n\rightarrow\mathbb{R}$ is characterized by a multilayer perceptron (MLP).
In particular, state-of-the-art neural field representations for 3D shape geometry has focused on optimizing for $f$ such that it approximates a signed distance field (SDF) representation of a shape, with its surface at isolevel 0 \cite{park2019deepsdf, sitzmann2020implicit, yifan2021iso, gropp2020implicit}.
High-fidelity reconstructions can be achieved from point observations in which complete object geometry has been observed; however, in real-world scanning scenarios, particularly for large scenes, only partial scans can be acquired, due to complex inter-scene occlusions and physical sensor limitations.

In such real-world scanning scenarios, crucial information from the camera pose trajectory -- that is, regions of space that are known to be empty as they were observed in front of a surface relative to the camera -- has been neglected in these reconstruction formulations.
We thus propose a new paradigm for 3D reconstruction with neural field representations, inspired by Poisson-based surface reconstruction techniques \cite{kazhdan2006poisson,kazhdan2013screened}.
To this end, we characterize surfaces optimized by neural fields as indicator functions, i.e., 0 outside and 1 inside the surface.
This is key to our indicator formulation for neural fields with empty space constraints, as values can be constrained to be 0; in contrast, with a distance field representation, empty space can be arbitrary values greater than zero.

In particular, optimization for the indicator function can be characterized by solving for the indicator function whose gradient inverts the normals of the point observations \cite{kazhdan2006poisson}.
The neural field formulation enables an efficient optimization of this variational problem, with an elegant way to include additional constraints for empty space regions.
In a series of experiments, we demonstrate that our indicator-based formulation for neural field representations for surface reconstruction produces more accurate and detailed reconstructions, improving $9.5\%$ in Chamfer distance over state of the art.

In summary, our contributions are:
\begin{itemize}
    \item We propose a new, Poisson-inspired indicator formulation for neural fields for high-fidelity surface reconstruction from point data.
    \item Our indicator formulation enables an elegant incorporation of intrinsic empty space information obtained from real-world scanning processes, for more accurate reconstruction.
\end{itemize}

\section{Related Work}

\paragraph{Implicit Neural Field Representations}

In recent years, neural field representations have been shown to have powerful capability in  modeling 3D shape surfaces.
Such neural networks implicitly model shape surfaces as level sets of a scalar-output multilayer perceptron that takes as input a coordinate in 3D space \cite{park2019deepsdf,mescheder2019occupancy,gropp2020implicit,chibane2020implicit,sitzmann2020implicit,yifan2021iso,xie2022neural}.
In particular, approximating a signed distance function (SDF) with an MLP has become a popular and effective approach for modeling 3D shapes.
DeepSDF~\cite{park2019deepsdf} approximated SDF representations of 3D shapes through auto-decoder training of an MLP, supervised by explicit computation of the SDF field of complete shape data.
Gropp et al.~\cite{gropp2020implicit} proposed to instead learn SDF approximations by applying an Eikonal loss on the gradient of the MLP, enabling optimization for shape reconstruction based only on observed points.
The sine-based MLP approach proposed by Siren~\cite{sitzmann2020implicit} further demonstrated the efficacy of sinusoid activations for SDF shape reconstruction leveraging an Eikonal loss.
Such MLP-based reconstruction of shape SDFs have also shown to effectively model deformable 3D shapes \cite{palafox2021npms,palafox2022spams}.

Such 3D shape modeling approaches, however, do not consider the extra information given by real-world scanning processes; given a sensor measuring surface points, we can infer that the line of space between the sensor and a surface measurement must be empty of geometry.
We thus instead propose to approximate an indicator function rather than an SDF, which enables a simple integration of free space information acquired in real-world scans.

In addition to SDF-based neural field representations for 3D shapes, Occupancy Networks (OccNet)~\cite{mescheder2019occupancy}  also proposed to approximate an occupancy function with an MLP, supervised with pre-computed point samples in the bounding volume of a complete shape.
The sampling-based approximation in its supervision tends to more easily lead to oversmoothed reconstruction results, in comparison to state-of-the-art SDF-based reconstruction.
Similar to the SDF-based approaches for neural fields, OccNet does not consider any empty space information in its shape reconstruction.
Our indicator-based formulation is instead inspired by the observation from Poisson Surface Reconstruction~\cite{kazhdan2006poisson}, that oriented point samples can be viewed as samples of the gradient of the indicator function representing the surface.

\paragraph{RGB-D Surface Reconstruction}

Reconstructing 3D surfaces from discrete range sensing measurements has had a well-studied history in computer graphics and computer vision.
Early work fit implicit functions to the point observations with radial basis functions \cite{muraki1991volumetric,carr2001reconstruction,turk2002modelling} or piecewise polynomial functions \cite{ohtake2005sparse,nagai2009smoothing}.
Various methods have also been developed to fit implicit indicator or signed distance field functions by formulating Laplacian systems on tetrahedral \cite{alliez2007voronoi} and octree representations \cite{calakli2011ssd}.

Poisson surface reconstruction~\cite{kazhdan2006poisson} was a seminal approach to surface reconstruction, which proposed to fit an implicit indicator function to an oriented point set, observing that the gradient of the indicator function must match the inward surface normals at surface points.
The optimization is formulated on an octree grid, with the variational problem  of approximating the vector field defined by the points transformed into a Poisson problem.
Screened poisson surface reconstruction~\cite{kazhdan2013screened} then extended the approach to incorporate constraints directly from the observed points.
Our approach is inspired these Poisson surface reconstruction approaches; however, rather than requiring a complex multigrid solver on an octree grid, we leverage the representation power of neural fields to formulate a simple reconstruction optimization that enables a clear incorporation of empty regions given during real-world scanning processes.

Volumetric grid-based reconstruction techniques have also been developed for 3D reconstruction, often incorporating empty space knowledge into aggregation of depth map observations into an SDF model of a scanned scene.
In particular, real-time 3D reconstruction methods leveraging voxel-based representations have leveraged empty space information from sensor observations in order to carve away spurious surfaces in known free regions \cite{newcombe2011kinectfusion,izadi2011kinectfusion,niessner2013hashing,dai2017bundlefusion}. 
Recent volumetric learning based approaches have also employed empty space information for shape and scene completion tasks~\cite{wu20153d,dai2017complete,song2017semantic,dai2018scancomplete,dai2020sgnn}.
We propose to lift this information away from explicit volumetric representations to a neural field representation not tied to any specific grid resolution, thus avoiding explicit surface point resampling or interpolation, and instead leveraging the implicit regularization properties in neural fields.

\section{Indicator-based 3D Neural Fields}

\begin{figure*}
	\centering
	\includegraphics[width=\linewidth]{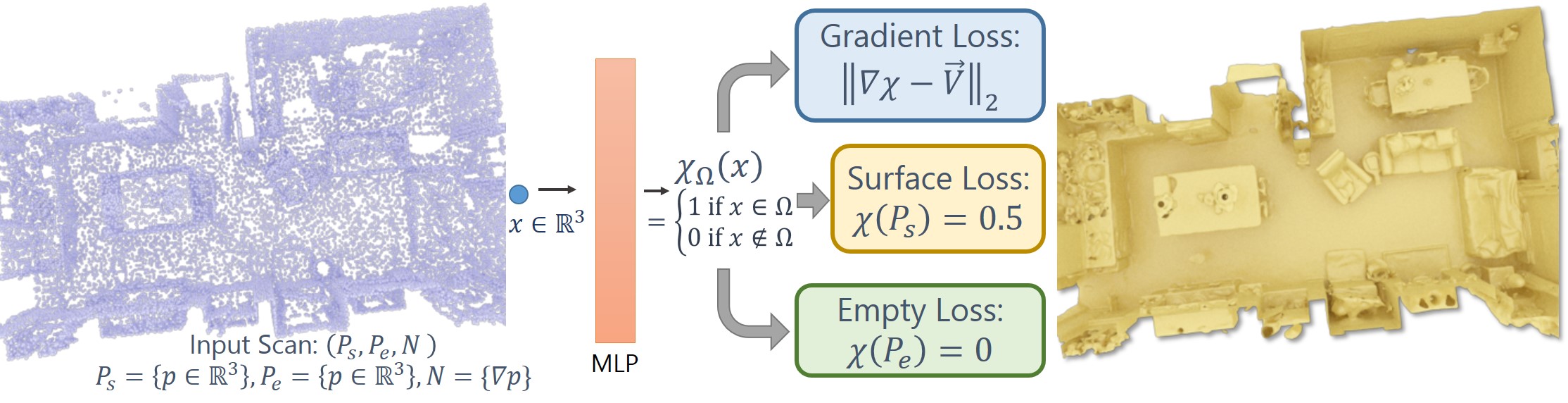}
	\vspace{-0.7cm}
	\caption{
	    From an input scan of surface observations $P_s$ taken from sensor locations $S$, we optimize for surface reconstruction with neural field $\chi$ approximating an indicator function.
	    From $P_s$, we can estimate the per-point orientations $N$ and sample empty space regions $P_e$; $\vec{V}$ is then estimated as the vector field defined by $P_s,N$.
	    We formulate surface reconstruction by using an MLP $\chi$ to approximate an indicator function. 
	    Oriented point observations constrain the surface and gradient of $\chi$, while empty space observations should be $0$-valued.
	    This produces an implicit, indicator-based surface reconstruction to accurately reflect input scan observations.
    }
	\label{fig:overview}
\end{figure*}

Given an input scan taken from sensor locations $S=\{s\in\mathbb{R}^3\}$ with measurements in the form of a point cloud $P_s = \{p\in \mathbb{R}^3\}$, with estimated (inward) normals $N=\{ \nabla p \in\mathbb{R}^3 \}$, our aim is to optimize for the neural network parameters $\theta$ of a neural field MLP $\chi(x,\theta) : \mathbb{R}^3\times\mathbb{R}^n\rightarrow \mathbb{R}$, such that $\chi$ approximates an indicator function representation of a surface $\delta\Omega$ which fits to $P_s,N$.

That is, the neural field $\chi$ takes as input a coordinate $x\in\mathbb{R}^3$, and applies MLP parameters $\theta$ to produce a scalar-valued output.
We then optimize for $\theta$ such that $\chi(x,\theta)$ is close to 1 when $x\in\Omega$ and 0 for $x\notin\Omega$. 
This enables a compact surface representation not tied to any explicit resolution, from which an explicit mesh representation can be easily extracted through Marching Cubes~\cite{lorensen1987marching}.

\begin{figure}[bp]
	\centering
	\includegraphics[width=0.8\linewidth]{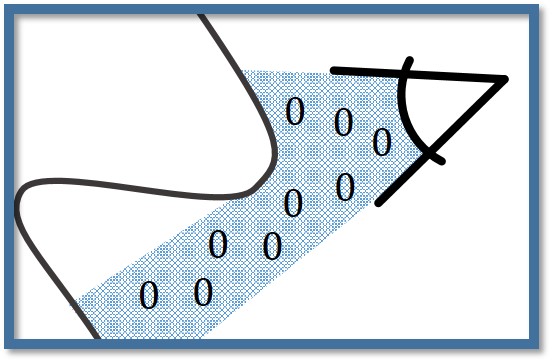}
	\vspace{-0.3cm}
	\caption{
	    A sensor observing a surface admits observations about empty space.
	    Regions between the sensor and the observed surface must be empty and thus 0-valued, as they precede a surface.
    }
	\label{fig:freespace}
\end{figure}

In particular, our indicator function formulation enables an elegant optimization formulation for surface reconstruction to unify constraints from input scan observations.
We observe that point cloud inputs are captured from sensors that are physically located in an environments; the sensor location in addition to its observed point measurements provides crucial information about empty space in the scene.
Empty space observations are visualized in Figure~\ref{fig:freespace}.
That is, the line segment between the sensor location $s$ and a point observation $p$ must be empty space, in order for an observation to have occurred no closer than $p$ from $s$.
We thus denote $P_e$ to represent points lying in such free space regions.
Given the binary nature of the indicator function representation, this enables us to formulate a novel empty space constraint for points in observed free space regions, $p\in P_e$, which should be $0$-valued.

In addition to our empty space constraint, we follow the observation from Poisson Surface Reconstruction~\cite{kazhdan2006poisson} that the gradient of the indicator function must be equal to the inward surface normal of the input points at the surface.
We thus also constrain the gradient of the estimated indicator function $\chi$ to fit the vector field defined by $P_s$.

Finally, we use the input point samples $P_s$ to encourage the isosurface of $\chi$  between 0 and 1 to pass through $P_s$.
The full loss formulation is discussed in Section~\ref{sec:opt}.

\section{Surface Reconstruction Optimization}\label{sec:opt}

We aim to optimize for an accurate surface reconstruction as an indicator function represented as a neural field $\chi(x, \theta)$, where $\chi$ is an MLP with parameters $\theta$ which takes as input $x\in\mathbb{R}^3$ and outputs the approximate indicator function.
We constrain $\chi$ and its gradient to match the input oriented point observations $P_s$, in addition to matching empty space given by the input scanning process.

Our loss is thus of the form:
\begin{align}
\begin{split}
    &L(P_s, P_e, \vec{V}, \theta) =\\ &\lambda_g L_g(P_s, \vec{V}, \theta) + \lambda_s L_s(P_s, \theta) + \lambda_e L_e(P_e, \theta),
\end{split}
\end{align}
where $L_g$ is a gradient constraint inspired by Poisson Surface Reconstruction~\cite{kazhdan2006poisson} to encourage $\chi$ to approximate an indicator function (Sec.~\ref{subsec:grad}), $L_s$ a surface constraint to encourage $\chi$ to interpolate input points (Sec.~\ref{subsec:surf}), and $L_e$ is an empty space constraint (Sec.~\ref{subsec:free}) to discourage $\chi$ from generating surface in empty regions.
$\lambda_g, \lambda_s, \lambda_e$ are scalar weighting factors to balance the loss terms.
Note that $\vec{V}$ represents the vector field defined by $P_s$, which we define in Section~\ref{subsec:gradientfield}.
This enables us to optimize $\chi$ to approximate an indicator function.

\subsection{Gradient Constraint} \label{subsec:grad}
The gradient of the estimated indicator function $\chi$ should match the vector field defined by the input point observations at the surface, giving
\begin{align}
    L_g(P_s, \vec{V}, \theta) = ||\nabla\chi(p) - \vec{V}(p)||_2^2,\quad p\in P_s,
\end{align}
where $\vec{V}$ is estimated by splatting normals computed from $P_s$ as described in Section~\ref{subsec:gradientfield}.
This inverting of the gradient operator relates to the Poisson Surface Reconstruction formulation~\cite{kazhdan2006poisson,kazhdan2013screened}; however, our use of stochastic gradient descent on an MLP enables solving for the above constraint without requiring any re-transformations.
This results in a very easy-to-implement optimization formulation, in contrast to re-transforming to a Poisson equation, discretization with the Galerkin method on a B-spline basis, and cascaded solve over a multigrid structure \cite{kazhdan2006poisson,kazhdan2013screened}.
In particular, our neural field formulation obviates the need for optimization over a grid, which is very complex in nature, requiring a multi-grid structure for capturing high resolutions, along with a linear~\cite{kazhdan2006poisson} or quadratic~\cite{kazhdan2013screened} interpolation kernel to facilitate gradient approximation.
In contrast, gradient computation of the MLP $\chi$ can be computed both simply and analytically, through the compute graph of the neural network representation.

\subsection{Surface Constraint}  \label{subsec:surf}
Similarly, $\chi$ should interpolate the input points $P$:
\begin{align}
    L_s(P_s, \theta) = ||\chi(p) - 0.5||_2^2,\quad p\in P_s
\end{align}
to encourage its isosurface between 0 and 1 to pass through the points in $P$.
That is, the surface should pass evenly between the 0/1 values at 0.5, so we penalize any deviation of $\chi$ from 0.5 at input observed point samples.
Similar to the gradient constraint, this loss is very simply applied to the MLP optimization.

\subsection{Empty Space Constraint} \label{subsec:free}
Finally, points in empty space should be 0-valued:
\begin{align}
    L_e(P_e, \theta) = ||\chi(p)||_2^2,\quad p\in P_e.
\end{align}
This enables incorporation of significant information given simply during the scanning process (as shown in Figure~\ref{fig:empty}) to guide surface reconstruction to avoid observed empty regions.
In particular, the indicator-based representation allows for easy characterization of space known to be free of surface geometry, as by definition it should be 0; in contrast, with a signed distance field representation, empty space regions are simply increasing in distance from the surface, which is more difficult to characterize.

\subsection{Normal and Gradient Field Estimation}\label{subsec:gradientfield}

We compute $N$ from $P_s$ by estimating local planes to the surface measurements.
When $P_s$ has been acquired as a sequence of depth frames, we leverage the grid structure of the frames to compute normals as a cross product of neighboring camera space positions of the depth values, following RGB-D scanning systems \cite{newcombe2011kinectfusion,niessner2013hashing,dai2017bundlefusion}.
When $P_s$ is captured as a point set, we estimate $N$ by locally fitting planes to the $k$-nearest neighbor points for each $p\in P_s$, given by the smallest-eigenvalued eigenvector of a principal component analysis. As normal direction in this scenario is ambiguous, we re-orient all normals to face the sensor location.
This provides a sparse set of normals $N$ corresponding to $P_s$.

For the gradient constraint, since a true indicator function is piecewise constant, its gradient would be unbounded at the surface $\delta\Omega$.
We thus instead consider convolving the indicator function with a smoothing filter, and use the gradient of the smoothed indicator function.
This is inspired by the Poisson Surface Reconstruction formulation~\cite{kazhdan2006poisson}, for which we refer to the proof that the gradient of the smoothed indicator function is equal to the vector field obtained by smoothing the normal field.

We then approximate the vector field $\vec{V}$ by applying a filter kernel function $F$ to the input oriented points.
This additionally enables us to estimate the values of $\vec{V}$ for regions near the input point observations, providing a degree of surface extrapolation under partial observations.

Then for a location $p\in\mathbb{R}^3$ near the surface points $P_s$, its normal is estimated by its $k$-nearest neighbors from $P_s$.
Normals of the $k$ nearest neighbors are clustered by greedy association starting with the closest oriented point, with each cluster normal represented as a gaussian-weighed average of the cluster element normals.
The closest cluster, based on distance to $p$, then provides its estimated normal as the cluster normal.
Empirically, we found this cluster-based approach to provide sharper reconstruction results than a gaussian-weighted average of the $k$ nearest neighbors.

\subsection{Empty Space Sampling}

To form our empty space constraint, we sample points in regions that are observed to be free based on $P_s$ and the sensor location(s) $S$.
Such sensor information is provided as part of the scanning process, in both commercial scanning products as well as with state-of-the-art RGB-D scanning systems \cite{whelan2015elasticfusion,dai2017bundlefusion}.
Thus, for a point $p\in P_s$ observed by a sensor at location $s$, we sample points $q$ along the ray $s+t\overrightarrow{(p-s)}$.
We sample 6 points per ray, of which 2 are sampled close to the observed surface points $(<0.02\textrm{m})$, to ensure that empty space is also constrained near the surface points.
The sampled points $q\in P_e$ are then subsampled to 1mm resolution, with a maximum of 4 million points.

\subsection{Implementation Details}

We represent $\chi$ as a 5-layer MLP with hidden dimension 256 and sine activation layers~\cite{sitzmann2020implicit}.
We use empirically determined loss weights $\lambda_g=1, \lambda_s=100$, and $\lambda_e=100$ for synthetic data and $\lambda_e=50$ for real scan data (due to noise in real scan data).
For optimization, we use an Adam optimizer with learning rate 0.0001, batch size of 100,000 points each from surface and empty samples, and optimize on a single NVIDIA GeForce GTX 1080 for 40 epochs.

We normalize all input point observations to $[-1,1]$, and zero-center all output predictions (i.e., instead of 0/1, we estimate a -0.5/0.5 indicator function, with loss constraints shifted accordingly), to better fit the outputs of a sine-based MLP.
Additionally, we use $k=20$ for gradient field estimation, and consider points near surface observations to be $<0.05$ from a point in $P_s$ after normalization into $[-1,1]$.

Meshes are extracted from the optimized $\chi$ by applying Marching Cubes~\cite{lorensen1987marching} at $640^3$ resolution samples of $\chi$.
\section{Experiments}

\begin{table}[tp]
\begin{center}
	\begin{tabular}{| l || c | c | c | c |}
		\hline
		Method &  CD ($\downarrow$) & IoU ($\uparrow$) & $\ell_2$ ($\downarrow$) \\  \thickhline
		SSD~\cite{calakli2011ssd} & 0.0189 & 0.504 & 2.610  \\ \hline
		SPSR~\cite{kazhdan2006poisson,kazhdan2013screened} & 0.0194 & 0.505 & 2.573  \\ \hline
        SPSR~\cite{kazhdan2006poisson,kazhdan2013screened} (+trim) & 0.0178 & 0.541 & 2.293  \\ \hline
		Siren~\cite{sitzmann2020implicit} & 0.0183 & 0.524 & 2.373  \\ \hline
        Ours & {\bf 0.0161} & {\bf 0.545} & {\bf 2.073}  \\ \hline
	\end{tabular}
	\caption{
	Evaluation of reconstruction from partial scans of 3D-FRONT~\cite{fu20213dfront} scenes.
	Our indicator-based neural field improves across all metrics in comparison with state-of-the-art traditional and neural network-based reconstruction methods.
	}
	\label{tab:3dfront}		
\end{center}
\end{table}

\begin{table}[bp]
\begin{center}
	\begin{tabular}{| l || c | c | c | c |}
		\hline
		Method &  CD ($\downarrow$) & IoU ($\uparrow$) & $\ell_2$ ($\downarrow$) \\  \thickhline
		SDF & 0.0170 & 0.521 & 2.173  \\ \hline
		SDF + High Off-Surface &  0.0170 & 0.518 & 2.165  \\ \hline
        Ours (No Empty) & 0.0167 & 0.537 & 2.132  \\ \hline
        Ours & {\bf 0.0161} & {\bf 0.545} & {\bf 2.073}  \\ \hline
	\end{tabular}
	\caption{
	Ablation study on reconstruction of partial scans of 3D-FRONT~\cite{fu20213dfront} scenes.
	Our indicator-based formulation improves over optimizing to approximated an SDF, along with optimizing with empty space constraints in a fully complementary fashion.
	}
	\label{tab:ablation}		
\end{center}
\end{table}

We demonstrate our approach on both synthetic and real scan data from 3D-FRONT~\cite{fu20213dfront} and ARKitScenes~\cite{arkitscenes}, respectively.
In the synthetic scenario, we use a virtual camera to scan depth sequence observations as input points, and evaluate against the ground truth meshes.
We use 100,000 points as input for scans of 3D-FRONT scenes, with point normals estimated from the respective virtually scanned depth maps as in Sec.~\ref{subsec:gradientfield}.
We evaluate 67 randomly sampled scenes from the 3D-FRONT dataset.
For real scan data, there are no ground truth meshes available, so we evaluate qualitatively.
For ARKitScenes scans, we use the Faro laser measurements sampled at 5mm resolution, with point normals estimated by PCA and re-oriented based on the scanner location(s), as in Sec.~\ref{subsec:gradientfield}.

\paragraph{Evaluation metrics}
In order to quantitatively evaluate reconstruction quality when ground truth meshes are available, we consider three measures: chamfer distance (CD), intersection over union (IoU), and $\ell_2$ distance.
For chamfer distance, 262,144 points are uniformly sampled from reconstructed and ground truth mesh surfaces, with all meshes normalized into the $[-1,1]$ range.
IoU and $\ell_2$ distance are computed over voxelization of the predicted and ground truth meshes at $128^3$, with IoU measured on the occupancy voxelization and $\ell_2$ as the $\ell_2$ distance of the distance field voxelizations (in voxel units), respectively.

\begin{figure*}[tp]
	\centering
	\includegraphics[width=0.97\linewidth]{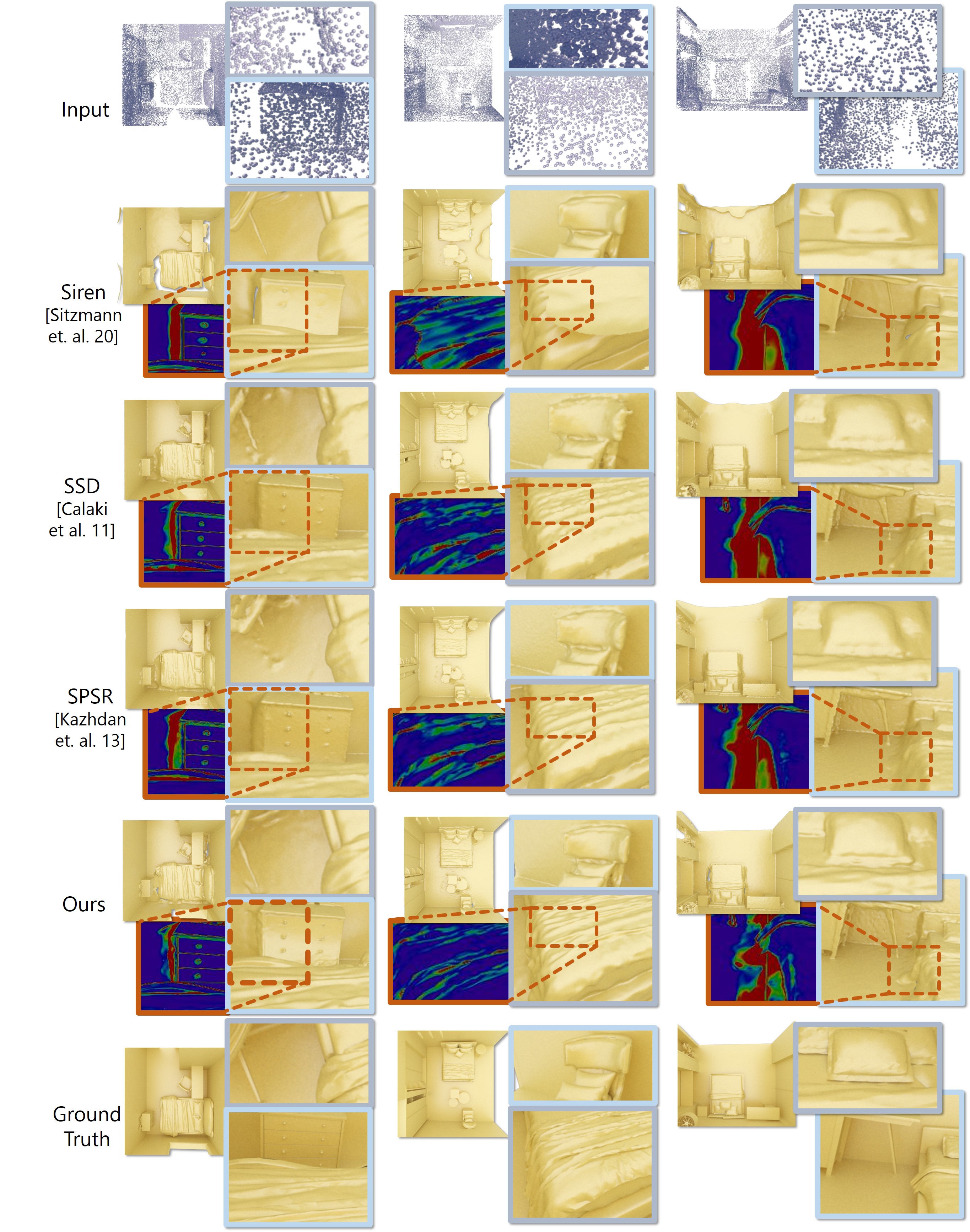}
	\vspace{-0.3cm}
	\caption{
	    Qualitative evaluation of surface reconstruction on 3D-FRONT~\cite{fu20213dfront} scans.
	    Error visualizations (blue 0, red 1) show normal deviation as cosine difference with ground truth in the zoom-in regions.
	    Our indictor-based neural field produces higher-fidelity details along with fewer spurious surfaces.
    }
	\label{fig:3dfront}
\end{figure*}

\begin{figure}[tp]
	\centering
	\includegraphics[width=0.9\linewidth]{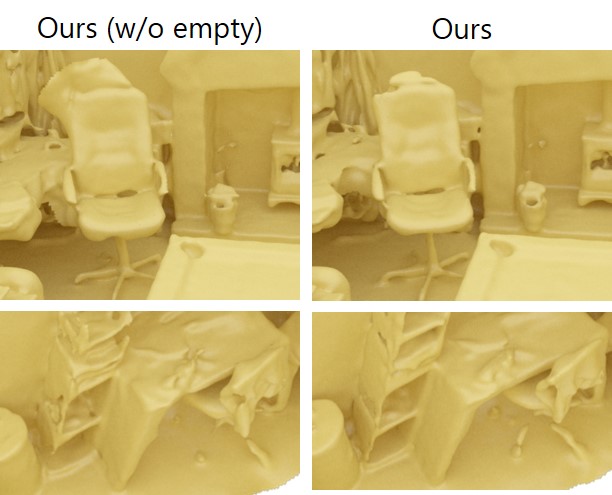}
	\caption{
	    Effect of empty space constraint on real-world ARKitScenes~\cite{arkitscenes} scans. Spurious surface generation is discouraged in observed empty regions, while fidelity is maintained near surface observations due to the near-surface empty space sampling. 
    }
	\label{fig:empty}
\end{figure}

\paragraph{Comparison with state of the art}
We compare with state-of-the-art classical and neural implicit surface reconstruction techniques: screened Poisson surface reconstruction (SPSR)~\cite{kazhdan2006poisson,kazhdan2013screened} optimizes for an indicator function by directly solving a Poisson equation formulation on an octree representation with the tailored adaptive multigrid solver proposed by the authors; SSD~\cite{calakli2011ssd} solves a similar variational formulation with an SDF representation; recently, Siren~\cite{sitzmann2020implicit} demonstrated state-of-the-art 3D surface reconstruction performance using a neural field representation, approximating an SDF.

Table~\ref{tab:3dfront} shows a comparison of our approach to these state-of-the-art methods for surface reconstruction for room-scale scans of 3D-FRONT~\cite{fu20213dfront} scenes. 
Note that since SPSR can generate spurious surfaces in regions of missing data, the official implementation also allows for post-processing trimming of the output reconstruction in regions of low input point density (\emph{+trim} in Table~\ref{tab:3dfront}), this can improve the measured reconstruction performance at the cost of removing surface extrapolation.
Our indicator-based formulation considering both surface and free space information from the input scans enables more accurate reconstruction than alternatives operating on octree representations or implicit SDF-based neural fields, which do not easily allow for empty space constraints.

We additionally show qualitative reconstruction results on 3D-FRONT scans in Figure~\ref{fig:3dfront}.
Our indicator-inspired formulation with empty space consideration enables reconstruction of finer-scale details (e.g., bed sheet wrinkles) while discouraging spurious surface ``ballooning'' in known free space (e.g., occluded side of the cabinet, left column).

\begin{figure*}[tp]
	\centering
	\vspace{-0.1cm}
	\includegraphics[width=0.95\linewidth]{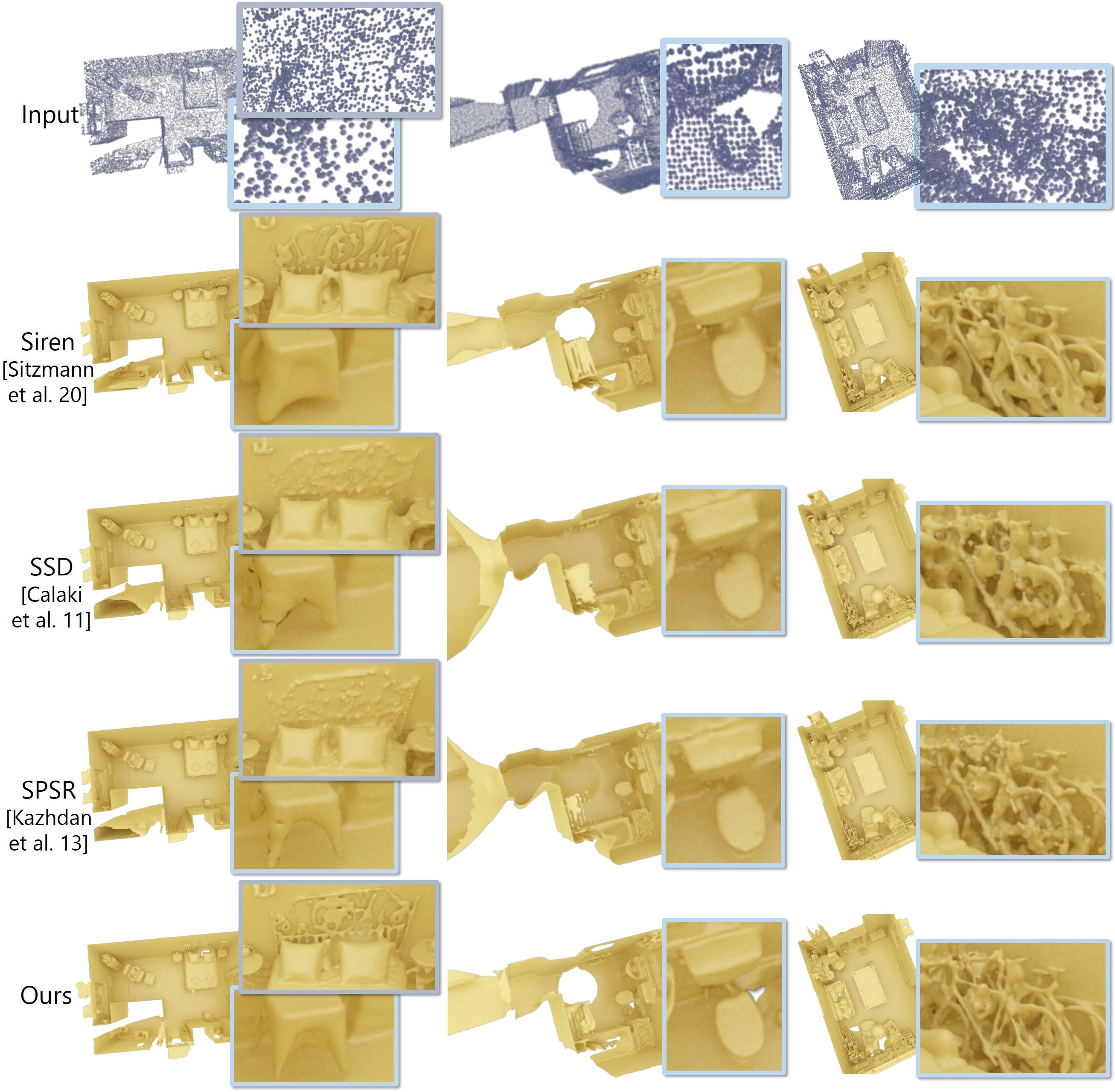}
	\vspace{-0.3cm}
	\caption{
	    Qualitative evaluation of surface reconstruction on real ARKitScenes~\cite{arkitscenes} scans. Our indicator-based neural field maintains finer-grained details (e.g., bed frame, toilet pipe, bicycles) while avoiding spurious generation in observed empty regions (e.g., stool, toilet seat).
    }
	\label{fig:arkit}
\end{figure*}

We further evaluate on 5mm-resolution scans from real-world captured ARKitScenes~\cite{arkitscenes} data in Figure~\ref{fig:arkit}.
As these scenes were captured in real environments, no ground truth meshes are available, and we show qualitative comparisons with state of the art.
Our approach to incorporate free space constraints with an indicator-guided neural field reconstruction enables more precise reconstruction of fine-grained details (e.g., bed frame, bicycles) while mitigating spurious surface generation (e.g., stool, toilet seat).

\begin{figure}[bp]
	\centering
	\includegraphics[width=0.9\linewidth]{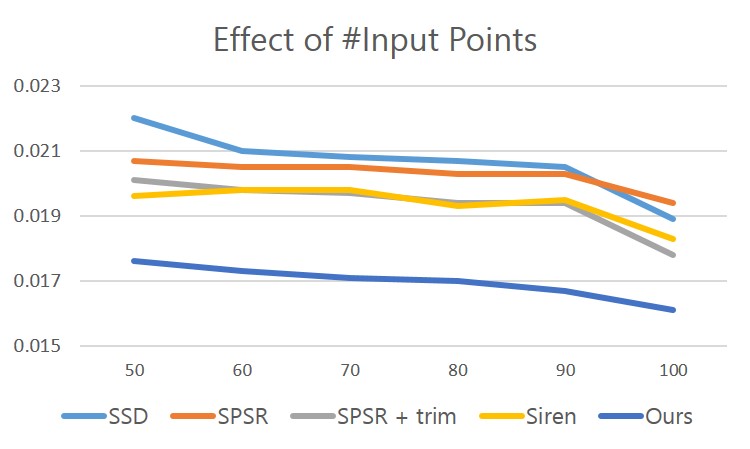}
	\caption{
	    Comparison against state of the art when varying the number of input points ($[50, 60, 70, 80, 90, 100]\times 10^3$). We measure Chamfer distance. Our indicator-based neural field consistently outperforms baselines, even under sparser point sampling.
    }
	\label{fig:numpoints}
\end{figure}

\paragraph{What is the impact of the indicator formulation?}

Table~\ref{tab:ablation} evaluates our approach when approximating a signed distance field output (using an Eikonal loss on the gradients to be 1) rather than an indicator function.
Reconstruction quality degrades when estimating an SDF, due to the greater complexity of the representation.
Additionally, the SDF representation does not easily admit an empty space constraint, as values off of the surface should be increasing with distance.
We can instead emulate an empty space constraint by encouraging non-surface values to be high, using loss constraint $L_{\textrm{sdf-free}} = \exp(100|\textrm{SDF}(p)|_1)$ for $p\notin P_s$.
This is less precise of a constraint, and does not strictly improve over using SDF only.
In contrast, our approach improves across all reconstruction quality measures when incorporating the empty space constraint.

\paragraph{What is the effect of the empty space constraint?}

Table~\ref{tab:ablation} shows that incorporating knowledge of free space given by the scanning process improves performance across all reconstruction metrics.
The effect is visualized in Figure~\ref{fig:empty}: our empty space constraint discourages spurious surface generation in observed empty regions, while maintaining high fidelity near surface observations due to the near-surface empty space sampling.

\paragraph{Effect of number of input points.}
We additionally analyze the effect of the number of input points samples per scene on 3D-FRONT~\cite{fu20213dfront} data in Figure~\ref{fig:numpoints}.
Under varying degrees of sparser input point sampling, our approach nonetheless still consistently outperforms all baselines.

\paragraph{Limitations and Future Work}
Our approach shows an effective incorporation of intrinsic scanning information for surface reconstruction; however, several limitations remain.
For instance, significant noise in the sensor pose estimation could lead to notable conflict between estimated empty space and surface regions.
In this context, it would be interesting to jointly solve for camera poses as part of the reconstruction process.
Additionally, while any surface extrapolation is discouraged to lie in observed free space, it will be pushed towards unobserved regions of space, which remains under-determined with respect to the true surface geometry.
Incorporation of a data-driven prior could improve general extrapolation behavior, for instance, learned based on the underlying semantics of the input point cloud.
\section{Conclusion}

We introduce a new indicator-based paradigm for 3D reconstruction with neural field representations.
In contrast to the conventional approximation for reconstruction with signed distance fields, our indicator representation admits a simple and effective surface reconstruction formulation to consider both surface observation constraints as well as empty space constraints, which are not easily represented in state-of-the-art alternatives.
This enables more accurate reconstruction in large-scale scanning scenarios.
We believe this perspective on the popular neural field representation for modeling 3D surfaces will provide new avenues in 3D reconstruction; e.g., when combining traditional reconstruction methods with learning-based techniques.

\section*{Acknowledgments}
This work was supported by the Bavarian State Ministry of Science and the Arts coordinated by the Bavarian Research Institute for Digital Transformation (bidt), the ERC Starting Grant Scan2CAD (804724), the German Research Foundation (DFG) Grant ``Making Machine Learning on Static and Dynamic 3D Data Practical'', and the German Research Foundation (DFG) Research Unit ``Learning and Simulation in Visual Computing.'' 

{\small
\bibliographystyle{ieee_fullname}
\bibliography{egbib}
}

\clearpage
\newpage
\begin{appendix}

\section{Additional Qualitative Results}
\begin{figure*}[tp]
	\centering
	\includegraphics[width=\linewidth]{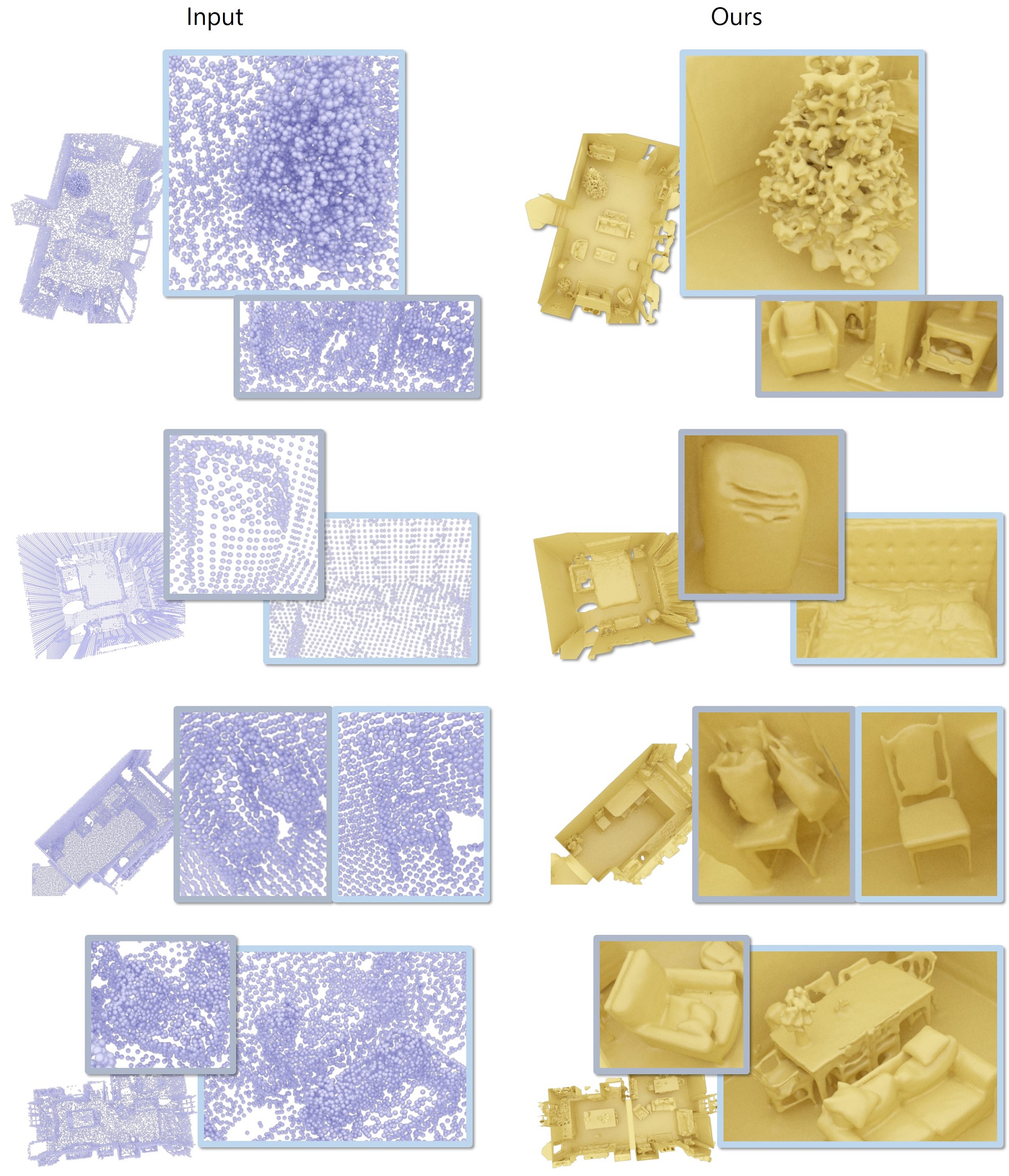}
	\caption{
	    Additional qualitative surface reconstruction results on real scans from ARKitScenes~\cite{arkitscenes}. Our indicator-based neural field achieves high fidelity reconstruction for finer-scale details (e.g., tree, chair legs) and global structures.
    }
	\label{fig:arkit_supp}
\end{figure*}

Figure~\ref{fig:arkit_supp} shows additional qualitative surface reconstruction results of our method on real-world ARKitScenes~\cite{arkitscenes} scans.
Our approach can effectively reconstruct smaller-scale structures such as tree in first row and the chair legs in bottom two rows.

\section{Additional Data Details}

To generate input data from the synthetic 3D scenes of the 3D-FRONT~\cite{fu20213dfront} dataset, we sample virtual cameras in each scene, approximately $1.5$ meters from each other, and facing frontwards, tilted upwards 30 degrees, and tilted downwards 30 degrees.
Depth maps are then extracted from rendered sensor images as input points. 
For each depth map, we use the grid structure of the depth image to compute normals; we compute camera space coordinates for each pixel using the camera intrinsic projection, and estimate normals as a cross product of the neighboring camera space positions.
This procedure follows state-of-the-art RGB-D scanning systems \cite{newcombe2011kinectfusion,niessner2013hashing,dai2017bundlefusion}.
Finally, all input points are randomly subsampled to 100,000 points per scene.

\section{Additional Baseline Details}

All methods were run on the same set of input points $P_s$ and estimated normals $\vec{N}$ used as input to our method.

\paragraph{Screened Poisson Surface Reconstruction (SPSR)~\cite{kazhdan2006poisson,kazhdan2013screened}} 
We used the authors' public implementation for SPSR\footnote{https://github.com/mkazhdan/PoissonRecon}.
SPSR was run at a depth of 9 for 3D-FRONT scans, and a depth of 10 for ARKitScenes scans, which we empirically found to perform the best for the respective data.
We follow with optional surface trimming from the authors' repository at a threshold of 6 (+trim in the main evaluation), which removes output geometry in regions of low input point density.

\paragraph{Smooth Signed Distance Surface Reconstruction (SSD)~\cite{calakli2011ssd}} 
We used the authors' public implementation for SSD\footnote{http://mesh.brown.edu/ssd/software.html}.
We run SSD at an octree maximum depth of 10 with weights $\lambda_0=10, \lambda_1=1, \lambda_2=1$ for the surface, gradient, and regularization energy terms, respectively, which we empirically found produce the best results for both datasets.

\paragraph{Siren~\cite{sitzmann2020implicit}} 
We used the authors' public implementation of Siren\footnote{https://github.com/vsitzmann/siren}.
We optimize for 18,000 epochs at a batch size of 100,000 points for a 5-layer MLP with 256 hidden units per layer.
We follow the learning rate and loss weights provided by the authors for the 3D reconstruction task.

\end{appendix}

\end{document}